\documentclass[10pt, conference, compsocconf]{IEEEtran}
\usepackage{amsmath}
\usepackage{algorithm}
\usepackage{algorithmic}
\usepackage{color}
\usepackage{supertabular,booktabs}
\usepackage{graphicx}
\usepackage{bm}

\begin{document}
\title{AlphaGomoku: An AlphaGo-based Gomoku Artificial Intelligence using Curriculum Learning}

\author
{\IEEEauthorblockN{Zheng Xie \IEEEauthorrefmark{1}\IEEEauthorrefmark{2},XingYu Fu\IEEEauthorrefmark{1},JinYuan Yu\IEEEauthorrefmark{1}
}

\IEEEauthorblockA
{
	\IEEEauthorrefmark{1}Likelihood Lab\\
}

\IEEEauthorblockA
{
	\IEEEauthorrefmark{2}Vthree.AI\\
}

$ $\\
$\{xiezh25,fuxy28,yujy25\}@mail2.sysu.edu.cn$
}
\maketitle

\begin{abstract}
In this project, we combine AlphaGo algorithm with Curriculum Learning to crack the game of Gomoku. Modifications like Double Networks Mechanism and Winning Value Decay are implemented to solve the intrinsic asymmetry and short-sight of Gomoku. Our final AI AlphaGomoku, through two days' training on a single GPU, has reached humans' playing level.
\end{abstract}

\begin{IEEEkeywords}
AlphaGo; Curriculum-Learning; Gomoku;
\end{IEEEkeywords}

\IEEEpeerreviewmaketitle

\section{Introduction}
Free style Gomoku is an interesting strategy board game with quite simple rules: two players alternatively place black and white stones on a board with 15 by 15 grids and winner is the one who first reach a line of consecutive five or more stones of his or her color. It is popular among students since it can be played simply with a piece of paper and a pencil to kill the boring class time. It is also popular among computer scientists since Gomoku is a natural playground for many artificial intelligence algorithms. Some powerful AIs are created in this field, to name a few, say YiXin, RenjuSolver. However, most of the AIs existed are rule-based requiring human experts to construct well-crafted evaluation function. To some extent, this kind of AIs are the slaves of humans' will without ideas formed by themselves. They are more like smart machines rather than intelligences.

Go is a far more complicate board game compared to Gomoku and cracking the game of Go is always the holy grail of the whole AI community. In the year of 2016, AlphaGo become the world first AI to defeat human Go grand master [3] and in 2017, it even mastered Go totally from scratch without human knowledge [1]. AlphaGo is, in general, a powerful universal solution to board games like Chess and Shogi [2]. The most exciting part of AlphaGo is that it masters how to play through \emph{learning} without any form of human-based evaluation.

Therefore, since the great potential of AlphaGo, we try to deploy the amazing algorithm in the game of Gomoku to construct artificial intelligence AlphaGomoku that can learn how to play free style Gomoku. However, customization is difficult since some intrinsic properties of free style Gomoku, e.g. asymmetry and short sight. It's easy to be stuck in poor local optimum where the white almost resigns and the black attacks blindly if we directly apply AlphaGo method.

Hence, to address these issues, we use the so-called Curriculum Learning [13] paradigm to smooth the training process. Modifications like Double Networks Mechanism and Winning Value Decay are implemented to alleviate the intrinsic issues of Gomoku. Through two days' training, AlphaGomoku has reached humans' performance.

All the source codes of this project are published in Github\footnote{https://github.com/PolyKen/15\_by\_15\_AlphaGomoku}.
\vspace{0.5cm}

\section{Original Components of AlphaGo}
\subsection{Bionics' Explanation of AlphaGo's Components}
AlphaGo's decision-making system is comprised of two units i.e. the policy-value network and the Monte Carlo Tree Search (MCTS) [6], both of which have some intuitive explanations in bionics.

Policy-value network functions like human's brain, which observes the current board and generates prior judgments, analogous to the human intuition, of the current situation. MCTS is similar to human's meditation or contemplation, which simulates multiple possible outcomes starting from the current state.

Like human's meditation process guided by intuition, the simulation procedure of MCTS is also controlled by the prior judgements generated from the policy-value network. Conversely, the policy-value network is also trained according to the simulation results from MCTS, an analogy that human's intellect is enhanced through deep meditation. 

The final playing decision is made by MCTS's simulation results, not the prior judgments from neural network, an analogy that rational person makes decision through deep meditation, not instant intuition. 

In the following subsections, we will make the above discussions more precise.

\subsection{Policy-Value Network}
The policy-value network $f_{\bm{\theta}}$, with $\bm{\theta}$ as the parameter, receives a tensor $\bm{s}$ characterizing the current board and outputs a prior policy probability distribution $\vec{\bm{p}}$ and value scaler $v$. Formally, we can write as:
\begin{equation}
(\vec{\bm{p}},v)=f_{\bm{\theta}}(\bm{s})
\end{equation}

Specifically, in AlphaGomoku, $\bm{s} = [\bm{X}, \bm{Y}, \bm{L}]$ is a $3$ by $15$ by $15$ tensor, where $15$ is the width and length of the board and $3$ is the number of channels. $\bm{X}$ and $\bm{Y}$ are consisted of binary values representing the presence of the current player's stones and opponent's stones respectively($X_i$ equals to one if the $i^{th}$ location is occupied by the current color stone; $X_i$ equals to zero if the $i^{th}$ location is either empty or occupied by the other color; $Y_i$'s assignment is analogous to $X_i$). $\bm{L}$ is the last-move channel whose values are binary in a way that $L_i=1$ if and only the last move the opponent takes is at the $i^{th}$ location. $\vec{\bm{p}} = (p_1, p_2, ..., p_{225})$ is a vector whose $i^{th}$ component $p_i$ represents the prior probability of placing the current stone to the $i^{th}$ location of the board. $v \in [-1, 1]$ represents the winning value of the current player, who is
about to place his or her stone at the current playing stage. The bigger $v$ is, the more network believes the current player is winning.

The network we used here is a deep convolutional neural network [9] equiped with residual blocks [8] since it can solve the degradation problem caused by the depth of neural network, which is essential for the learning capability. Other mechanisms like Batch-Normolization [10] are used to further improve the performance of our policy-value net. The architecture of the net is consisted of three parts: 
\begin{itemize}
\item \emph{Residual Tower}: Receives the raw board tensor and conducts high-level feature extraction. The output of residual tower is passed to policy head and value head separately.
\item \emph{Policy Head}: Generates the prior policy probability distribution vector $\vec{\bm{p}}$.
\item \emph{Value Head}: Generates the winning value scalar $v$.
\end{itemize}
And a detailed topology is shown in Table 1, Table 2, and Table 3.
\begin{table}[!ht]
	\centering
	\begin{tabular}{ll}
		\hline
		\textbf{Layer}\\
		\hline
		(1)  Convolution of 32 filters size 3 with stride 1\\
		(2)  Batch-Normalization \\
		(3)  Relu Activation \\
		(4)  Convolution of 32 filters size 3 with stride 1\\
		(5)  Batch-Normalization \\
		(6)  Relu Activation \\
		(7)  Convolution of 32 filters size 3 with stride 1\\
		(8)  Batch-Normalization \\
		(9)  Shortcut \\
		(10) Relu Activation \\
		(11) Convolution of 32 filters size 3 with stride 1\\
		(12) Batch-Normalization \\
		(13) Relu Activation \\
		(14) Convolution of 32 filters size 3 with stride 1\\
		(15) Batch-Normalization \\
		(16) Shortcut \\
		(17) Relu Activation \\
		\hline\\
	\end{tabular}
	\caption{Residual Tower}\label{Table 1}
\end{table}

\begin{table}[!ht]
	\centering
	\begin{tabular}{ll}
		\hline
		\textbf{Layer} \\
		\hline
		(1)  Convolution of 2 filters size 1 with stride 1\\
		(2)  Batch-Normalization \\
		(3)  Relu Activation \\
		(4)  Flatten \\
		(5)  Dense Layer with $225$ dim vector output \\
		(6)  Softmax Activation \\
		\hline\\
	\end{tabular}
	\caption{Policy Head}\label{Table 2}
\end{table}

\begin{table}[!ht]
	\centering
	\begin{tabular}{ll}
		\hline
		\textbf{Layer} \\
		\hline
		 (1)  Convolution of 1 filter size 1 with stride 1 \\
		 (2)  Batch-Normalization \\
		 (3)  Relu Activation \\
		 (4)  Flatten \\
		 (5)  Dense Layer with $32$ dim vector output \\
		 (6)  Relu Activation \\
		 (7)  Dense Layer with scaler output \\
		 (8)  Tanh Activation \\
		\hline\\
	\end{tabular}
	\caption{Value Head}\label{Table 3}
\end{table}

\subsection{Monte Carlo Tree Search}
MCTS $\alpha_{\bm{\theta}}$, instructed by policy-value net $f_{\bm{\theta}}$, receives the tensor $\bm{s}$ of current board and outputs a policy probability distribution vector $\vec{\bm{\pi}}=(\pi_1, \pi_2, ..., \pi_{225})$ generated from
multiple simulations. Formally, we can write as:
\begin{equation}
\vec{\bm{\pi}}=\alpha_{\bm{\theta}}(\bm{s})
\end{equation}
Compared with those of $\vec{\bm{p}}$, the informations of $\vec{\bm{\pi}}$ are more well-thought since MCTS chews the cud of current situation deliberately. Hence, $\vec{\bm{\pi}}$ serves as the ultimate guidance for decision-making in AlphaGo algorithm. There are three types of policies used in our AlphaGomoku program based on $\vec{\bm{\pi}}$:
\begin{itemize}
\item \emph{Stochastic Policy}: AI chooses action randomly with respect to $\vec{\bm{\pi}}$, i.e.
\begin{equation}
a \sim \vec{\bm{\pi}}
\end{equation}
\item \emph{Deterministic Policy}: AI plays the current optimal move, i.e.
\begin{equation}
a \in \mathop{\arg\max}_{i}\vec{\bm{\pi}}
\end{equation}
\item \emph{Semi-Stochastic Policy}: At the every beginning of the game, the AI will play stochastically with respect to policy distribution $\vec{\bm{\pi}}$. And as the game continunes, after a user-specified stage, the AI will adopt deterministic policy.
\end{itemize}
and they are used in different contexts which we shall cover later.

Unlike the black-box property of policy-value network, we know exactly the logic inside MCTS algorithm. For a search tree, each node represents a board situation and the edges from that node represent all possible moves the player can take in this situation. Each edge stores the following statistics:
\begin{itemize}
\item \emph{Visit Count, $N$}: the number of visits of the edge. Larger $N$ implies MCTS is more interested in this move. Indeed, the policy probability distribution $\vec{\bm{\pi}}$ of state $\bm{s}$ is derived from the visit counts $N(\bm{s},k)$ of all edges from $\bm{s}$ in a way that:
\begin{equation}
\pi_i=N(\bm{s},i)^{\frac{1}{\tau}}/\sum_{k=1}^{225}N(\bm{s},k)^{\frac{1}{\tau}}
\end{equation} 
where $\tau$ is the temperature parameter controlling the trade-off between exploration and exploitation. If $\tau$ is rather small, then the exponential operation will magnify the differences between components of $\vec{\bm{\pi}}$ and therefore reduces the level of exploration.
\item \emph{Prior Probability, $P$}: the prior policy probability generated by the network after network evaluates the edge's root state $\bm{s}$. Larger $P$ indicates the network prefers this move and hence may guide MCTS to exam it carefully. While note that large $P$ does not guarantee a nice move since it is merely a prior judgment, an analogy to human's instant intuition.
\item \emph{Mean Action Value, $Q$}: the mean of the values of all nodes of the subtree under this edge and it represents the average wining value of this move. We can write as:
\begin{equation}
Q(\bm{s},i)=\frac{1}{N(\bm{s},i)} \sum_{v \in A}\pm v 
\label{equ 6}
\end{equation}
, where $A=\{v|(\_,v)=f_{\bm{\theta}}(\bm{\hat{s}})$, for $\bm{\hat{s}}$ in the subtree of $\bm{s} \}$. The tricky part here is the plus-minus sign "$\pm$". $v$ is the wining value of the current player of $\bm{\hat{s}}$, not necessarily $\bm{s}$'s current player, while $Q$ here is to evaluate the wining chance of the current player of $\bm{s}$ if he or her chooses this move and therefore we need to adjust $v$'s sign accordingly.
\item \emph{Total Action Value, $W$}: the total sum of the values of all nodes of the subtree under this edge. $W$ serves as an intermediate variable when we try to update $Q$ since we have the following relation: $Q = W/N$. 
\end{itemize}
Before each action is played, MCTS will simulate possible outcomes starting from the current board situation $\bm{s}$ for multiple times. Each simulation is made up of the following three steps:
\begin{itemize}
\item \emph{Selection}: Starting from the root board $\bm{s}$, MCTS iteratively select edge $j$ such that: 
\begin{equation}
j \in \mathop{\arg\max}_{k}\{Q(\bm{\hat{s}},k)+U(\bm{\hat{s}},k)\}
\end{equation}
at each board situation $\bm{\hat{s}}$ under $\bm{s}$ till $\bm{\hat{s}}$ is a leaf node. The expression $Q(\bm{\hat{s}},j)+U(\bm{\hat{s}},j)$ is called the upper confidence bound, where:
\begin{equation}
U(\bm{\hat{s}},j)=c_{puct}P(\bm{\hat{s}},j)\frac{\sqrt{\sum_{k}N(\bm{\hat{s}},k)}}{1+N(\bm{\hat{s}},j)}
\end{equation}
and $c_{puct}$ is a constant controlling the level of exploration. The design of upper confidence bound is to balance relations among mean action value $Q$, prior probability $P$ and visit count $N$. MCTS tends to select the moves with large mean action value, large prior probability and small visit count. 
\item \emph{Evaluation and Expansion}: Once MCTS encounters a leaf node, say $\tilde{s}$, which haven't been evaluated by network before, we let $f_{\bm{\theta}}$ outputs its prior policy probability distribution $\vec{\bm{p}}$ and wining value $v$. Then, we create $\tilde{s}$'s edges and initialize their statistics as: $N(\tilde{s},i)=0, Q(\tilde{s},i)=0, W(\tilde{s},i)=0, P(\tilde{s},i)=p_i$ for the $i^{th}$ edge.
\item \emph{Backup}: Once we finish the evaluation and expansion procedure, we traverse reversely along the path to the root and update statistics of all the edges we pass in the backup procedure in a way that $N=N+1, W=W\pm v, Q=W/N$. The plus-minus sign "$\pm$" here is to implement Equ 6.
\end{itemize}
Fig. 1 demonstrates the pipeline of decision-making process of AlphaGo algorithm.
\begin{figure}[ht]
	\centering
	\includegraphics[scale=0.26]{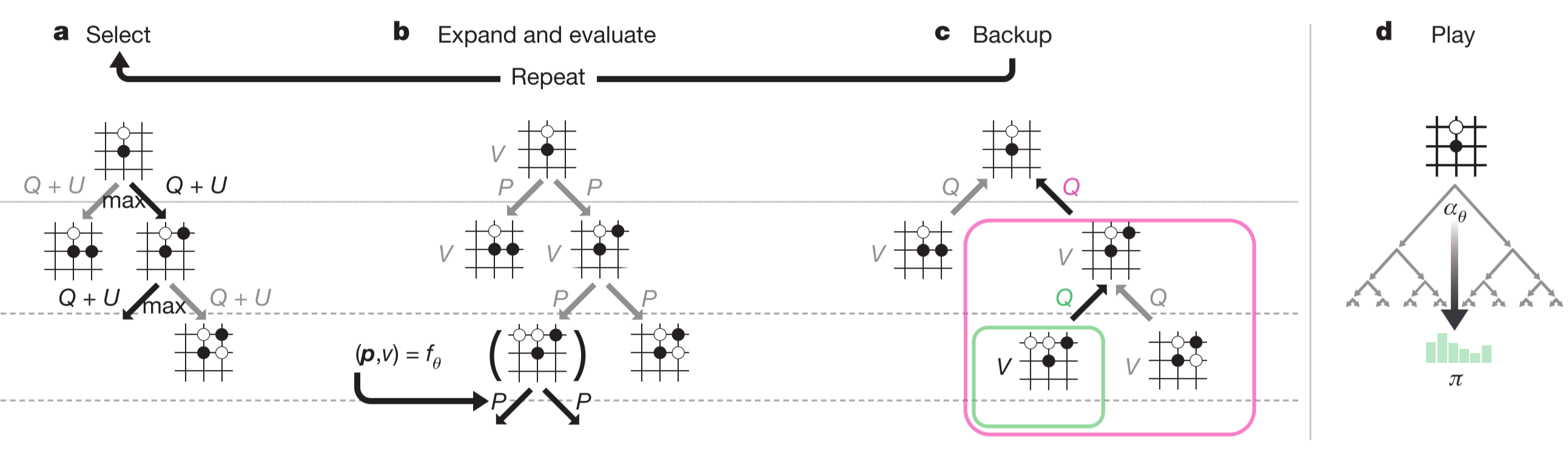}
	\caption{Decision-Making of AlphaGo Algorithm [1]} \label{fig 1}
\end{figure}

One tricky issue to note is the problem of \emph{end node}, e.g. draw or win or lose, which doesn't have legal children nodes. If end node is encountered in simulation, instead of executing the evaluation and expansion procedure, we directly run the backup mechanism, where the backup value is -1 if it is a win(lose) node or 0 if it is a draw node.

\subsection{Training Target of Policy-Value Network}
The key to the success of AlphaGo algorithm is that the prior policy probability distribution $\vec{\bm{p}}$ and wining value $v$ provided by policy-value network narrow MCTS to a much smaller search space with more promising moves. The guidance of $\vec{\bm{p}}$ and $v$ directly influences the quality of moves chose by MCTS and therefore we need to train our network in a wise way to improve its predictive intellect. There are three criteria set to a powerful policy-value network:
\begin{itemize}
\item It can predict the \emph{winner} correctly.
\item The \emph{policy distribution} it provides is similar to a deliberate one, say the one simulated by MCTS.
\item It can \emph{generalize} nicely.
\end{itemize}
To these ends, we apply Stochastic Gradient Descend algorithm with Momentum [11] to minimize the following loss function:
\begin{equation}
L(\bm{\theta}) = (z-v)^2-\vec{\bm{\pi}}^Tlog\vec{\bm{p}}+c\Vert \bm{\theta} \Vert^2
\end{equation}
where $c$ is a parameter controlling the $L_2$ penalty and $z$ is the result of the whole game, i.e.
\begin{equation}
z=\left\{\begin{array}{rcl}
1 && \text{if current player wins.}\\
0 && \text{if game draws.}\\
-1 && \text{if current player loses.}
\end{array}\right.
\end{equation}
And we can see the three terms in the loss function reflecting the three criteria that we mentioned above.

\subsection{Multi-Threading Simulation}
The time bottleneck of the decision-making process is the simulation of MCTS, which can be accelerated greatly using the multi-threading technique.

There are three points needed to be addressed to implement the asynchronous simulation:
\begin{itemize}
\item \emph{Expansion Conflict}: If two threadings happen to encounter the same leaf node and expand the node simultaneously, then the number of children nodes under this node will be mistaken and a conflict occurs. To address this issue, we maintain an expanding list which is a list of leaf nodes under expansion. When a threading encounters a leaf node, instead of expanding the node instantly, the threading checks the expanding list to see if the current node is in it. If yes, then the threading waits for a short period of time to let the other threading expands the node and keeps on selecting after the other threading finishes its expansion. If no, then just executes the normal expansion procedure.
\item \emph{Virtual Loss}: To let different threadings try as various paths as possible, after each selection, we discredits the selected edge virtually by increasing its $N$ and deceasing its $W$ temporarily to deceive other threadings into choosing other edges. And the DeepMind terms the amount of increasing and decreasing as virtual loss. Clearly, each threading needs to clear out the virtual loss in the backup procedure.
\item \emph{Dilution Problem}: If the virtual loss and the number of threadings are set too high, then the simulation numbers of the most promising moves may be diluted severely since many threadings are deceived to search other less promising edges. Therefore, the tunning of related hyperparameters is crucial.
\end{itemize}
\vspace{0.5cm}

\section{Customize AlphaGo to the Game of Gomoku}
\subsection{Asymmetry and Short Sight of Free Style Gomoku}
Although AlphaGo is, in general, a universal board game algorithm which also successfully master the game of Chess and Shogi besides Go, it is still difficult to make the algorithm work in Gomoku without proper customization. As a matter of fact, our first attempt of directly applying AlphaGo algorithm to Gomoku without customization fails, where the AI is stuck in poor local optimum.

The reasons for the difficulty lie in the fact that free style Gomoku is extremely asymmetric and short-sighted compared to Go, Chess, and Shogi.

\emph{Asymmetry}: The black player, the side placing stone first, has greater advantage than the white player, which results in unbalanced training dataset, i.e. black wins far more games than white. Training on such dataset directly will easily corrupt the value branch of the network, making the white AI almost resign and the black AI arrogant since they both mistakenly agree that the black is winning regardless of the current circumstance. Besides, since the asymmetry of the game, the strategies of the black and the white are quite different, where the black is prone to attack while the white tends to defend. It's hard to master such divergence with a single network. As we observe, if we apply the original AlphaGo algorithm to Gomoku, the white AI will sometimes attack blindly indicating the white strategy is negatively influenced by the black strategy.

\emph{Short Sight}: Unlike Go, Chess, and Shogi where the global situation of the board determines the winning probability, Gomoku is more short-sighted where local recent situation is more important than global long-term situation. Therefore, AI should decrease the winning value back propagated from the simulation of future, which is not implemented in the original AlphaGo algorithm.

To alleviate the issues, we propose the following modifications to customize the AlphaGo algorithm to free style Gomoku: Double Networks Mechanism and Winning Value Decay, which shall be clear in the below subsections.

\subsection{Double Networks Mechanism}
To solve the problem of asymmetry, we construct two policy-value networks to learn black and white strategies separately. Specifically, the black net is trained solely on the black moves and the white net is trained solely on the white moves. In simulation, the black net or white net are applied depending on the color of the current expanding node to give prior policy distribution and predicted winning value of the node.

By implementing the Double Networks Mechanism, we can see significant improvement of AlphaGomoku's performance since it now plays asymmetrically, which fits the traits of the game.

\subsection{Winning Value Decay}
There are two kinds of backward processes happens in the original AlphaGo algorithm, one is the backup stage of simulation where the predicted winning value $v$ of the leaf node is used to update the $Q$ values of the nodes in the current simulation path and another is the labelling process of $z$ where each move's $z$ is set according to the final result of the game. To solve the problem of short sight, we let the values to decay exponentially in the above processes and the resulted improvement is promising since the AI now focuses more on the recent events, which are dominantly significant in free style Gomoku.
\vspace{0.5cm}

\section{Curriculum Learning}
\subsection{Intuition of Curriculum Learning}
Curriculum Learning [13], a machine learning paradigm proposed by Bengio et.al, mimics the way human receive education. It introduces relatively easy concepts to the learning algorithm at the initial training stage and gradually increases the difficulty of the learning mission. By training like this, the learning algorithm can take advantage of previously learned basic concepts to ease the learning of more high-level abstractions. Bengio et.al have shown empirically that curriculum learning can accelerate the convergence of non-convex training and improve the quality of the local optimum obtained.

Back to the case of Gomoku, although AlphaGo algorithm is capable of learning the Go strategy without the guidance of human knowledge, it's computationally intractable for most organizations to conduct such learning and the quality of the AI obtained cannot be guaranteed. Hence, we train AlphaGomoku in a curriculum learning pipeline to accelerate the training and secure the performance of the AI.

Specifically, our training pipeline can be divided into three phases: Learn Basic Rule, Imitate Mentor AI, and Self-play Reinforcement Learning. In the first two phases, AlphaGomoku learns basic strategy and receives guidance from the mentor AI. In the final phase, AlphaGomoku learns from its own playing experiences and enhance its playing performance. We can summarize the training pipeline using a famous Chinese saying: \emph{The master teaches the trade, but apprentice's skill is self-made}.

Note that in all phases we need to train the black network and white network separately on corresponding moves. Fig. 2 demonstrates the general pipeline of our curriculum learning.

\begin{figure}[ht]
	\centering
	\includegraphics[scale=0.35]{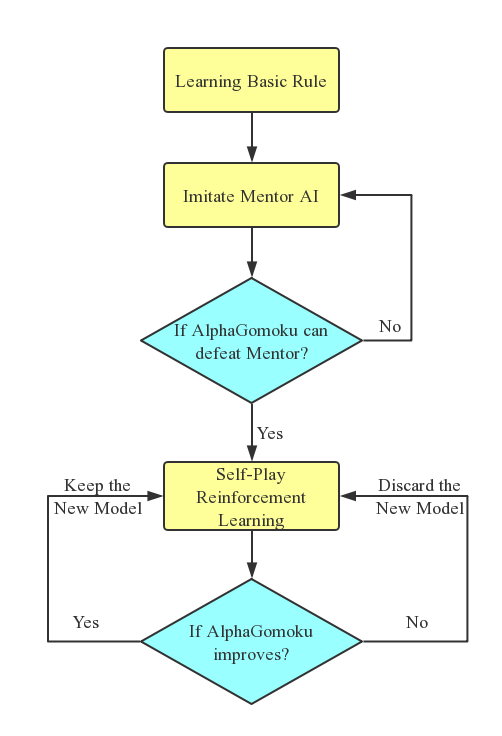}
	\caption{Curriculum Learning}\label{fig 2}
\end{figure}

\subsection{Learn Basic Rule}
At the very beginning of training, we teach the basic strategy of Gomoku to AlphaGomoku. To be specific, we randomly generate eighty thousand basic moves and let the networks learn the moves using mini-batch stochastic gradient descent with momentum. The moves generated include basic attack and basic defense.

For an attack instance, we randomly generate $\{\bm{s},\vec{\bm{\pi}},z\}$. $\bm{s}$ is a situation, whose current player is black, containing a consecutive line of four black stones without being blocked in both sides. $\vec{\bm{\pi}}$, the target policy distribution, is a one hot vector whose items are zero except the place which leads black to five. $z$, the target winning value, is set to one.

For a defense instance, we generate $\{\bm{s},\vec{\bm{\pi}},z\}$. $s$ is a situation, whose current player is white, containing a consecutive line of four black stones being blocked in only one side. $\vec{\bm{\pi}}$, the target policy distribution, is a one hot vector whose items are zero except the place which prevents the black from reaching five. $z$, the target winning value, is set to zero.

\subsection{Imitate Mentor Gomoku Artificial Intelligence}
We implement a rule-based tree search Gomoku AI as the mentor of AlphaGomoku. The mentor AI competes against itself and AlphaGomoku learns the games using mini-batch stochastic gradient descent with momentum. As we can observe, after learning the from mentor AI, AlphaGomoku has formed some advanced strategies like "three-three", "four-four" and "three-four" and its playing style is quite similar to mentor AI.

When AlphaGomoku can successfully defeat mentor AI with a great margin, we stop the imitation process to avoid over-fitting and further enhance the performance of AlphaGomoku through self-play reinforcement learning.

\subsection{Self-Play Reinforcement Learning}
After the imitation, we train the networks through self-play reinforcement learning, where we maintain a single search tree guided by double policy-value networks and let it compete with itself using semi-stochastic policy. After each move, the search tree alters its root node to the move it takes and discards the remainder of the tree until the game ends. We collect all the data generated in several games and sample uniformly from the collected data to train the networks. The structure of each training data is:
\begin{equation}
\{\bm{s},\vec{\bm{\pi}},z\}
\end{equation}
Note that Gomoku is invariant under rotation and reflection, and hence we can augment training data by rotating and reflecting the board using seven different ways. After each training, we evaluate the training effect by letting the trained network compete with the currently strongest model using semi-stochastic policy. If the trained model wins, then we set the newly trained model to be the currently strongest and discard the old model. If the trained model loses, then we discard the newly trained model and keep the old one to be the currently strongest. Fig. 3 demonstrates the pipeline of self-play reinforcement learning.

We discuss three important questions that worth special attention in the above paragraph:
\begin{itemize}
\item \emph{Why we need the evaluation procedure?} To avoid poor local optimum by discarding badly performed trained model.
\item \emph{Why we adopt semi-stochastic policy in self-play and evaluation?} Firstly, we add randomness into our policy to conduct exploration since more possible moves will be tried. Secondly, we let our model to behave discreetly after a certain stage to avoid bad quality data.
\end{itemize}

Another interesting observation to note is the variation of time spent in each self-play game. The time will first decay and then extend. The reason for this phenomenon is that as the agent evolves across time, it first grasps the attacking technique, which shortens the game, and then learns the defending technique, which prolongs the game.
\begin{figure}[ht]
	\centering
	\includegraphics[scale=0.35]{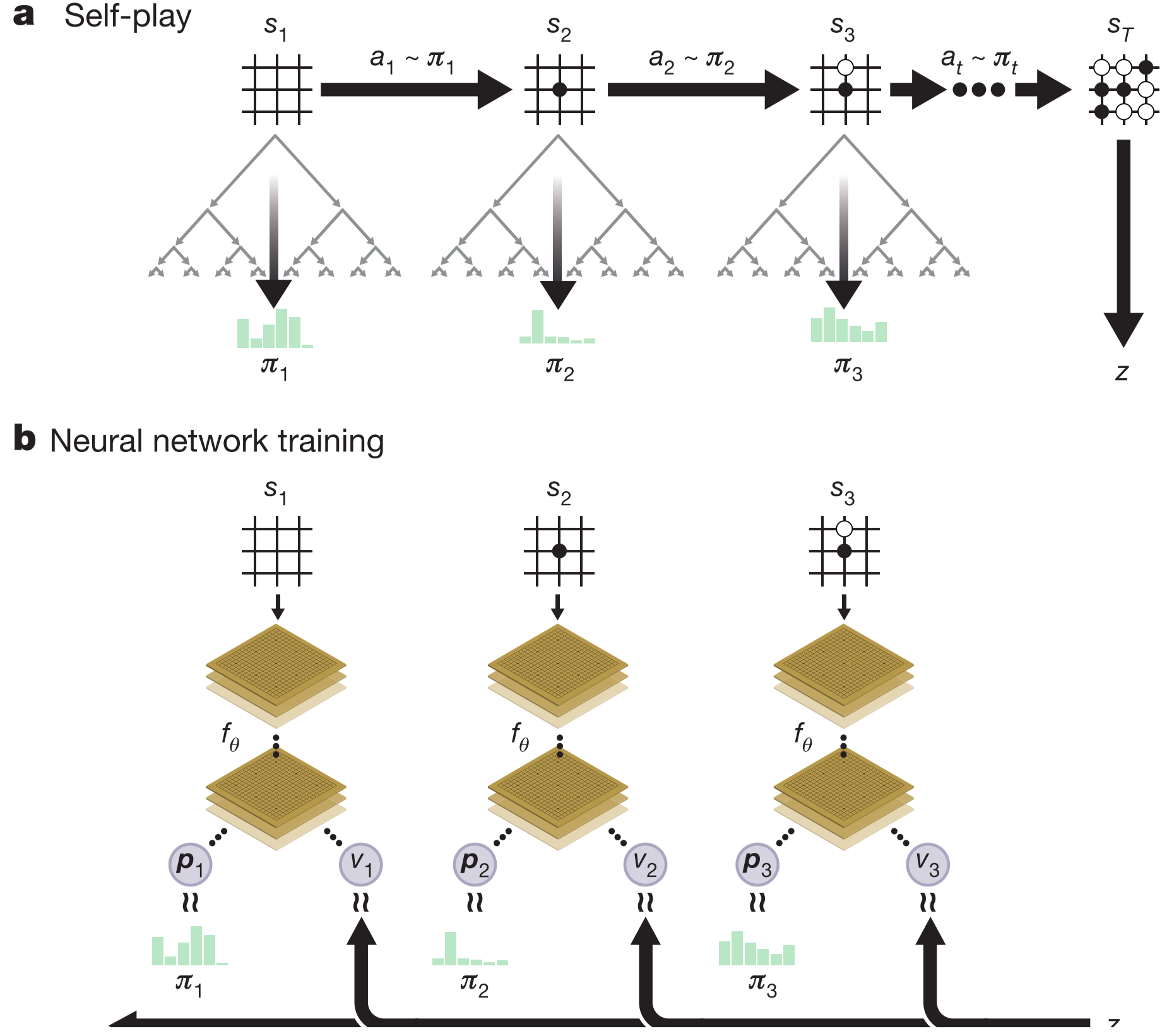}
	\caption{Pipeline of Self-Play Reinforcement Learning [1]}\label{fig 3}
\end{figure}
\vspace{0.5cm}

\section{EXPERIMENT}
\subsection{Mentor AI vs AlphaGomoku}
We let mentor AI compete with version-14 AlphaGomoku for one hundred games, in which AlphaGomoku plays black and white for fifty games respectively. We can see clearly that AlphaGomoku dominates the game and surpasses mentor AI. Fig. 4 shows the statistics of the competition. Fig. 5 and Fig. 6 are two sample games between mentor and AlphaGomoku.
\begin{figure}[ht]
	\centering
	\includegraphics[scale=0.35]{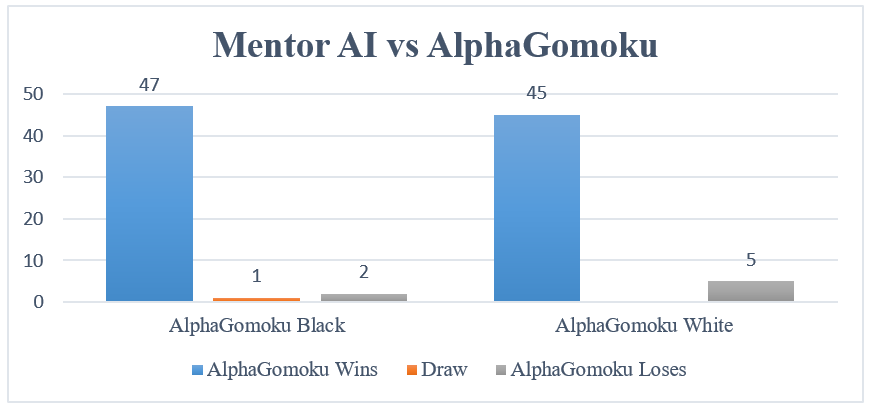}
	\caption{Statistics of Mentor vs AlphaGomoku} \label{fig 4}
\end{figure}
\begin{figure}[ht]
	\centering
	\includegraphics[scale=0.35]{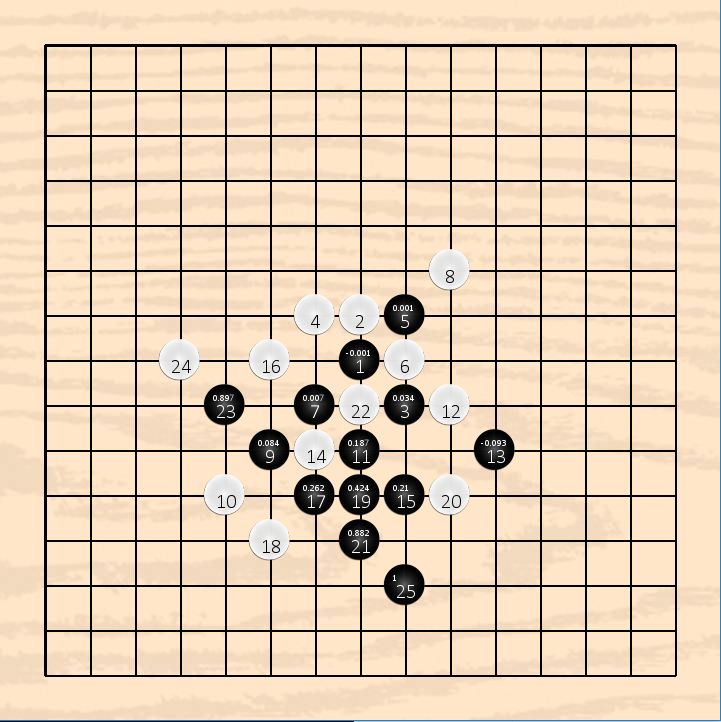}
	\caption{Mentor(White) vs AlphaGomoku(Black)} \label{fig 5}
\end{figure}
\begin{figure}[ht]
	\centering
	\includegraphics[scale=0.35]{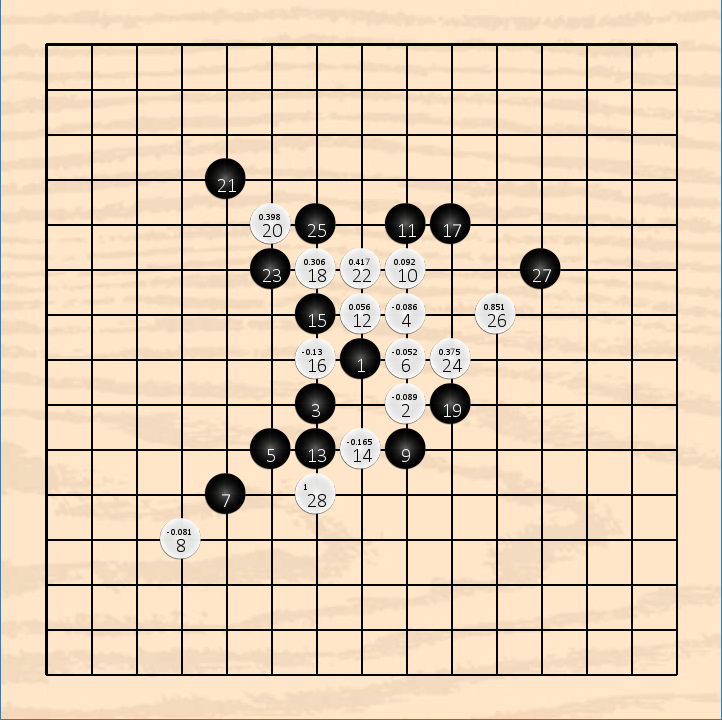}
	\caption{Mentor(Black) vs AlphaGomoku(White)} \label{fig 6}
\end{figure}

\subsection{Human vs AlphaGomoku}
Version-14 AlphaGomoku use wechat mini program happy Gomoku to challenge random online players. Fig. 7 shows the statistics of our online test. Note that two of the games of the online test are played by Macau Gomoku professional and AlphaGomoku only loses to him when it plays black stone. Fig. 8, Fig. 9, Fig. 10 and Fig. 11 are the sample games of human vs AlphaGomoku.
\begin{figure}[ht]
	\centering
	\includegraphics[scale=0.35]{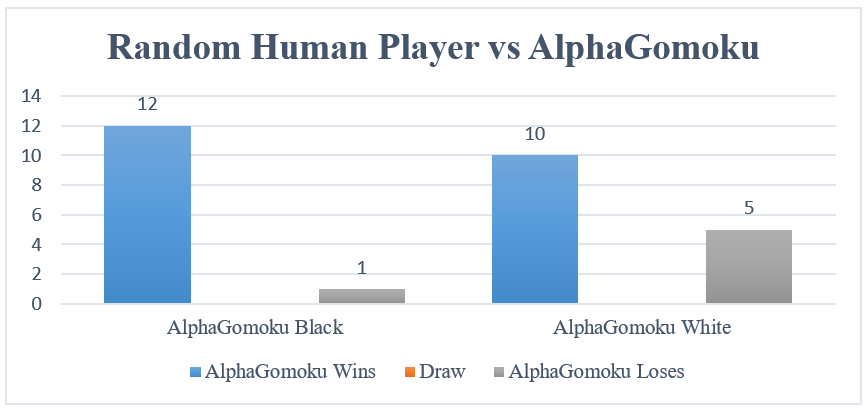}
	\caption{Statistics of Human vs AlphaGomoku} \label{fig 7}
\end{figure}
\begin{figure}[ht]
	\centering
	\includegraphics[scale=0.35]{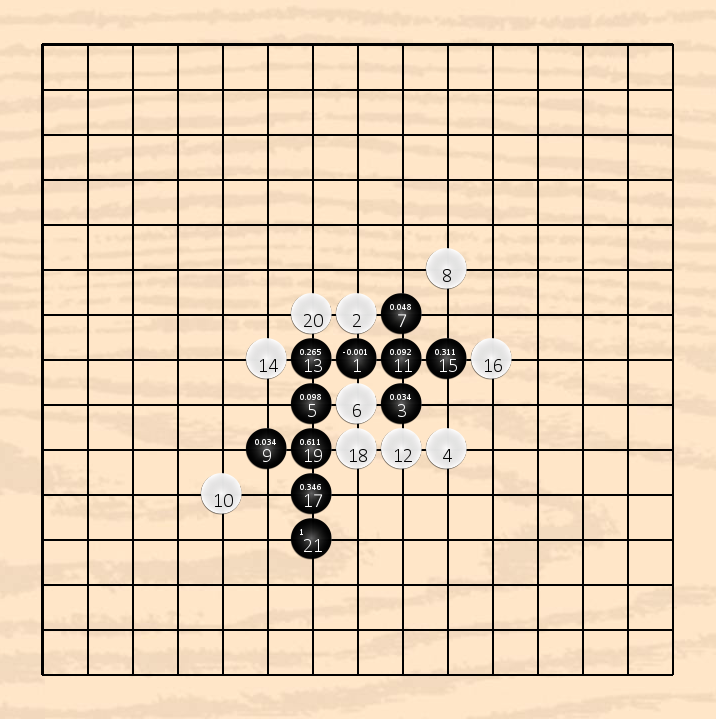}
	\caption{Human(White) vs AlphaGomoku(Black)} \label{fig 8}
\end{figure}
\begin{figure}[ht]
	\centering
	\includegraphics[scale=0.35]{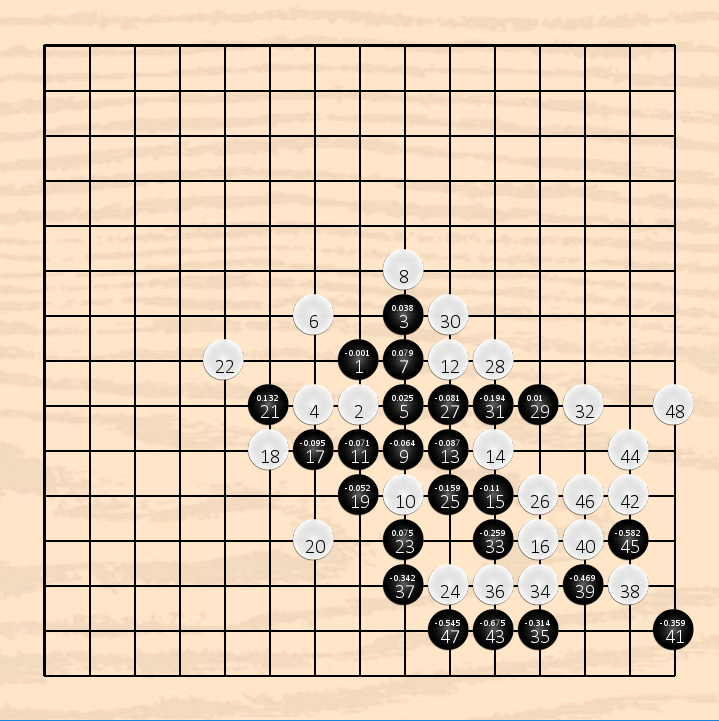}
	\caption{Human(White) vs AlphaGomoku(Black)} \label{fig 9}
\end{figure}
\begin{figure}[ht]
	\centering
	\includegraphics[scale=0.35]{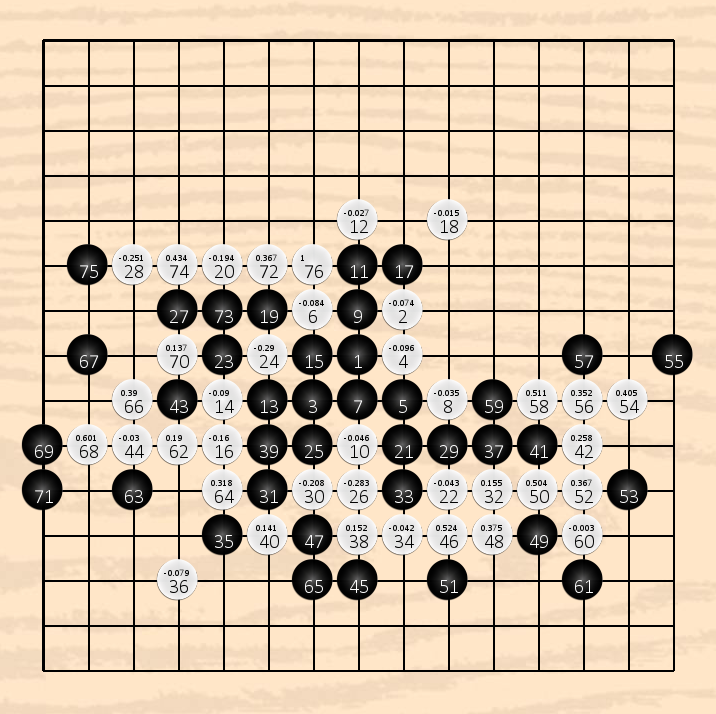}
	\caption{Human(Black) vs AlphaGomoku(White)} \label{fig 10}
\end{figure}
\begin{figure}[ht]
	\centering
	\includegraphics[scale=0.35]{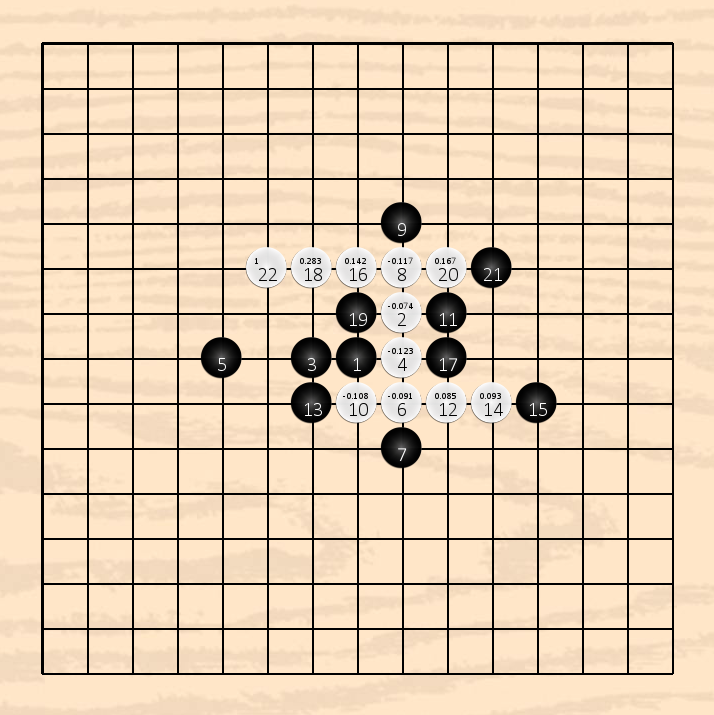}
	\caption{Human(Black) vs AlphaGomoku(White)} \label{fig 11}
\end{figure}
\vspace{0.5cm}

\section*{Acknowledgment}
We would like to say thanks to BaiAn Chen from Vthree.AI and MingWen Liu from ShiningMidas Private Fund for their generous help throughout the research. We are also grateful to ZhiPeng Liang and Hao Chen from Sun Yat-sen University for their supports of the training process of our Gomoku AI. Without their supports, it's hard for us to finish such a complicated task.


\begin{thebibliography}{1}
\bibitem{IEEEhowto:kopka}
Silver D, Schrittwieser J, Simonyan K, et al. Mastering the game of go without human knowledge[J]. Nature, 2017, 550(7676): 354.\label{ref 1}

\bibitem{IEEEhowto:kopka}
Silver D, Hubert T, Schrittwieser J, et al. Mastering chess and shogi by self-play with a general reinforcement learning algorithm[J]. arXiv
preprint arXiv:1712.01815, 2017.\label{ref 2}

\bibitem{IEEEhowto:kopka}
Silver D, Huang A, Maddison C J, et al. Mastering the game of Go with deep neural networks and tree search[J]. nature, 2016, 529(7587): 484.\label{ref 3}

\bibitem{IEEEhowto:kopka}
Mnih V, Kavukcuoglu K, Silver D, et al. Human-level control through deep reinforcement learning[J]. Nature, 2015, 518(7540): 529. \label{ref 4}

\bibitem{IEEEhowto:kopka}
Sutton R S, Barto A G. Reinforcement learning: An introduction[M]. MIT press, 1998. \label{ref 5}

\bibitem{IEEEhowto:kopka}
Browne C B, Powley E, Whitehouse D
, et al. A survey of monte carlo tree search methods[J]. IEEE Transactions on Computational Intelligence
and AI in games, 2012, 4(1): 1-43. \label{ref 6}

\bibitem{IEEEhowto:kopka}
Goodfellow I, Bengio Y, Courville A, et al. Deep learning[M]. Cambridge: MIT press, 2016.\label{ref 7}

\bibitem{IEEEhowto:kopka}
He K, Zhang X, Ren S, et al. Deep residual learning for image recognition[C]//Proceedings of the IEEE conference on computer vision and
pattern recognition. 2016: 770-778. \label{ref 8}

\bibitem{IEEEhowto:kopka}
Krizhevsky A, Sutskever I, Hinton G E. Imagenet classification with deep convolutional neural networks[C]//Advances in neural information processing systems. 2012: 1097-1105.\label{ref 9}

\bibitem{IEEEhowto:kopka}
Ioffe S, Szegedy C. Batch normalization: Accelerating deep network training by reducing internal covariate shift[J]. arXiv preprint arXiv:1502.03167, 2015.\label{ref 10}

\bibitem{IEEEhowto:kopka}
Ruder S. An overview of gradient descent optimization algorithms[J]. arXiv preprint arXiv:1609.04747, 2016.\label{ref 11}

\bibitem{IEEEhowto:kopka}
Knuth D E, Moore R W. An analysis of alpha-beta pruning[J]. Artificial intelligence, 1975, 6(4): 293-326. \label{ref 12}

\bibitem{IEEEhowto:kopka}
Bengio, Yoshua, et al. "Curriculum learning."Proceedings of the 26th annual international conference on machine learning. ACM, 2009. \label{ref 13}

\end{thebibliography}
\end{document}